\documentclass[sigconf]{acmart}
\usepackage{multirow}
\usepackage{xcolor}
\usepackage{tcolorbox}
\usepackage{hyperref}
\usepackage{caption}  
\usepackage{booktabs}

\AtBeginDocument{%
  }

\setcopyright{acmlicensed}
\copyrightyear{2025}
\acmYear{2025}
\acmDOI{ https://dbpmworkshop.github.io/files/personaldata.pdf}

\acmConference[LLM-DPM '2025]{Next Gen Data and Process
Management: Large Language Models and Beyond}{June 22, 2025}{Berlin, Germany}

\acmISBN{978-1-4503-XXXX-X/2018/06}

\begin{document}

\title{Detection of Personal Data in Structured Datasets Using a Large Language Model}

\author{Albert Agisha Ntwali}
\email{albert.agishantwali@hs-aalen.de}
\affiliation{%
  \institution{Aalen University of Applied Sciences}
  \city{Aalen}
  \country{Germnany}
}

\orcid{0000-0001-6194-250X}

\author{Luca Rück}
\email{luca.rueck@studmail.htw-aalen.de}
\affiliation{%
  \institution{Aalen University of Applied Sciences}
  \city{Aalen}
  \country{Germany}}

\author{Martin Heckmann}
\email{martin.heckmann@hs-aalen.de}
\affiliation{%
  \institution{Aalen University of Applied Sciences}
  \city{Aalen}
  \country{Germany}}

\renewcommand{\shortauthors}{Agisha N. et al.}

\begin{abstract}

We propose a novel approach for detecting personal data in structured datasets, leveraging GPT-4o, a state-of-the-art Large Language Model.
A key innovation of our method is the incorporation of contextual information: in addition to a feature's name and values, we utilize information from other feature names within the dataset as well as the dataset description.
We compare our approach to alternative methods, including Microsoft Presidio and CASSED, evaluating them on multiple datasets:
DeSSI, a large synthetic dataset, datasets we collected from Kaggle and OpenML as well as MIMIC-Demo-Ext, a real-world dataset containing patient information from critical care units.

Our findings reveal that detection performance varies significantly depending on the dataset used for evaluation. CASSED excels on DeSSI, the dataset on which it was trained.
Performance on the medical dataset MIMIC-Demo-Ext is comparable across all models, with our GPT-4o-based approach clearly outperforming the others.
Notably, personal data detection in the Kaggle and OpenML datasets appears to benefit from contextual information. This is evidenced by the poor performance of CASSED and Presidio (both of which do not utilize the context of the dataset) compared to the strong results of our GPT-4o-based approach.

We conclude that further progress in this field would greatly benefit from the availability of more real-world datasets containing personal information.

\end{abstract}



\maketitle

\section{Introduction}
Recent years have seen a strong increase in the amount of data created and stored in a digital form. 
In many cases also personal data is collected.
However, this vast accumulation of data and its digital availability brings with it substantial challenges, particularly concerning compliance with data protection regulations \cite{EuropeanUnion, GDPRReg, CAG}.

At the heart of these regulations is the General Data Protection Regulation (GDPR) of the European Union, which is widely considered the gold standard for data privacy \cite{EuropeanUnion, GDPRReg}. The GDPR not only aims to protect individuals' fundamental rights regarding the processing of their data but also imposes strict penalties on organizations that fail to comply, with potential fines reaching up to 4\% of global annual turnover \cite{gdprlocalfines}. The significance of the GDPR has led to its adoption as a foundational reference for personal data protection worldwide, influencing similar regulations in various countries \cite{Greenleaf2023global, Moschovitis:2021}.

To comply with these regulations and for ethical reasons, organizations need to implement effective measures to detect and manage personal data. 
In light of the often large volume of data, this requires powerful automated detection tools \cite{Fugkeaw2021, Presidio}.

\paragraph{Problem Statement}
Detecting personal data within structured datasets poses unique challenges. The format of the document being analyzed significantly influences the effectiveness of personal data detection efforts. The GDPR highlights the importance of context in defining personal data, indicating that an individual can often be identified indirectly through a combination of seemingly innocuous information. For example, while an individual's age alone may not be identifying, when combined with other attributes, such as their organization or job title, it can lead to their identification\cite{GDPRReg}.
Existing solutions tend not to integrate information across several columns and thereby neglect important information \cite{Kuzina2023}. 
For instance, a "Device Number" column in an IT asset database typically represents a device's serial number and may not initially appear to contain personal information. However, if the database is linked to an employee database that tracks assigned hardware, the device number could serve as an identifier for an employee, making it personal data.
Without proper contextual comprehension, these solutions misclassify data and result in compliance issues and privacy violations.

\paragraph{Objective of the Study}

We aim to develop an effective system for detecting personal data in structured datasets by integrating contextual information into the detection process. We propose a novel approach that employs large language models (LLMs), specifically the GPT-4o model, to improve detection accuracy. By comparing this model with established benchmarks, including Microsoft Presidio and CASSED models, we seek to highlight the advantages of context-aware detection methods.

Furthermore, we target an evaluation of these models on realistic real-world data. Previously used synthetic datasets bear the risk of not reflecting the real challenges properly and, hence, leading to wrong conclusions.
For this reason, we collected a total of 33 datasets from Kaggle and OpenML and also included MIMIC-Demo-Ext, a real-world dataset containing patient information from critical care units, in our evaluation.

\paragraph{Contributions of this Paper}
To summarize, the contributions of this paper are as follows:

\begin{enumerate}
\item \textbf{Novel Methodology for Personal Data Detection:} We introduce a novel approach that leverages the capabilities of Large Language Models (LLMs), specifically GPT-4o, to enhance the detection of personal data in structured datasets. By integrating contextual information into the detection process, we aim to improve accuracy and adaptability compared to traditional methods.

\item \textbf{Comprehensive Benchmarking:} We compare GPT-4o with established models like Microsoft Presidio and CASSED. The evaluation utilizes the DeSSI, Kaggle, OpenML, and MIMIC-Demo-Ext datasets to evaluate the detection of personal data. Comparing the performance across diverse datasets allows a better assessment of the strengths and limitations of each detection system.

\item \textbf{Real-world vs. Synthetic Data:} A key element of our evaluation is the comparison of results obtained on synthetic vs. real-world data. With this, we aim to determine if synthetic data can serve as a reliable benchmark.
\end{enumerate}

\section{Related Work}

The field of personal data detection has gained increasing attention, especially in light of evolving privacy regulations. However, before moving forward, it is essential to define what we mean by \textit{personal data}.

\subsection{Definition of Personal Data}\label{sec:defin_pers}

The concept of personal data includes several related terms. The most commonly used are:
\begin{enumerate}
    \item\label{item:pers} Personal Data: According to Article 4 of the GDPR, personal data is defined as “any information relating to an identified or identifiable natural person” \cite{finckthey1}. Examples include information about a person’s physical properties, contact details or identification numbers. Personal data can be divided into two categories: directly identifiable information and indirectly identifiable information.
    \item Personally Identifiable Information (PII): PII is defined as any information that can directly identify a person without the need for additional information, such as a bank account number or an email address\cite{Kulkarni2021}.
    \item Person-related Data: Person-related data cannot directly identify a person but relates to a natural person and may lead to identification when combined with other data \cite{GDPRReg}. Information such as gender or age does not directly identify a person but can be used in combination with other data for identification.
\end{enumerate}
When we refer to personal data, we use the definition of the GDPR (\ref{item:pers}) as a supergroup of PII and person-related data.
We specifically used this definition when we labeled data as personal or non-personal in the datasets we collected.
Some authors also use the term \textit{Sensitive Data} synonymously with personal data (e.g. \cite{Kuzina2023}) even though the GDPR defines it as a subgroup of personal data\cite{EuropeanUnion}.

\subsection{Detection of Personal Data from Unstructured Data}

Most approaches focus on the detection of personal data in unstructured data (e.g. text) \cite{shahriar_identifying_2024, Yang2023, Kulkarni2021}.

Initial approaches for personal data detection in unstructured data primarily utilized traditional named entity recognition (NER) techniques, which were adapted from broader natural language processing (NLP) applications, including sensitive information detection and PII detection \cite{Yang2023, Park2020sensitive}. 
Very recent approaches also use Large Language Models (LLMs) for this task \cite{Yang2023}.

\subsection{Detection of Personal Data from Structured Data}
Research on personal data detection from structured data is much more scarce.
One prominent example is Microsoft Presidio\cite{Presidio} which was designed for general entity recognition tasks yet applied to PII detection.
It is based on a set of recognizers for predefined classes \cite{presidioEntities} (compare Tab:\ref{tab:presidio_identifiers}).
Presidio employs a combination of approaches including NER, regular expressions, rule-based logic, checksums, and machine learning to identify sensitive data. Presidio also offers options for connecting to external PII detection models and supports customization in PII identification and anonymization. While effective in many scenarios, these approaches can be less effective in handling context-dependent variations of personal data \cite{Liu2014}.

More recent models use advanced machine learning methods to integrate information from a column's name with the values of different cells in this column.
Examples are Sherlock \cite{Hulsebos20191},  which employs deep neural networks, and SIMON \cite{Azunre2019}, which relies on a character-level neural network and an LSTM architecture.

The advent of language models like BERT \cite{devlin-etal-2019-bert}
allowed for more robust approaches to personal data detection.
BERT transforms sequences of text into vector embeddings which are able to capture the semantics of the text much better than previous approaches. 
As a consequence, variations in a feature's name or its values have a much less detrimental effect.
The CASSED model is based on such an approach \cite{Kuzina2023}.
More precisely, it employs DistilBERT \cite{Sanh2020}, a lightweight variant of BERT, to convert a feature's name and some exemplar feature values into an embedding. 
For doing so, the text sequence of the feature's name and the values of the feature are concatenated into a string, separated by delimiters, allowing DistilBERT to treat the column as a quasi-natural sentence.
The maximum token length of 512 elements of DistilBERT limited the amount of information that could be used. 
For this reason, CASSED is not able to include additional information from the neighboring columns.
In parallel to this DistilBERT path, a rule-based path is used to identify personal data.
Both paths are then combined via sigmoid functions to convert the scores for all 20 different classes of personal and non-personal data categories CASSED can recognize into probabilities (compare Tab:~\ref{tab:dessi_personal_nonpersonal} for a list of the classes). These probabilities are then used to make a decision.
CASSED was trained using DeSSI \cite{DeSSI}, a large dataset annotated with the aforementioned 20 classes.

\section{Methodology}

In the following, we will first introduce the different datasets we use to evaluate the different approaches.
Next, we will describe our GPT-4o-based approach in detail.
We will then explain which models we used as benchmarks for our own approach and how they needed to be adapted to be suitable for this comparison.
After that, we will also describe the performance metrics we used to benchmark the models.

Our implementation is open-source and publicly available on GitHub\footnote{The implementation can be accessed at: \url{https://github.com/agishaalbert/personal-data-detection-LLMs/}}.

\subsection{Dataset Selection}

Our evaluation relies on a large range of datasets containing personal information.
The first and by far largest dataset we use is the DeSSI dataset (Dataset for Structured Sensitive Information) \cite{Kuzina2023, DeSSI}, created to simulate real-world relational database challenges. The authors give no exact definition what they consider \textit{sensitive information}, yet, they reference the GDPR in their work and the list of classes they use (compare Tab:~\ref{tab:dessi_personal_nonpersonal}) aligns with the definition of personal data of the GDPR \cite{Kuzina2023}.
From this, we conclude that they also use the definition of the GDPR as we do. DeSSI consists of over 31,000 database columns with 100 rows each, derived from open-source datasets (e.g., Kaggle), synthetic data generated via Python libraries like Faker, and pseudo-anonymized real-world data\cite{FakerCommunity, Kuzina2023}. Columns were intentionally designed with randomized or misleading headers to reflect real-world inconsistencies and avoid reliance on misleading column headers. The dataset was randomly split in ratios of 60/20/20 percent among
training/validation/test datasets. For training the CASSED model, we use the training and validation part, and for the evaluation of CASSED, Presidio, and our GPT-4o-based approach, we only use the 6272 columns of the test set. For our experiments, we mapped the original 20 semantic classes into two categories: personal and non-personal data (compare Tab:~\ref{tab:dessi_personal_nonpersonal} in the appendix). 
Additionally, we extracted 13 datasets from Kaggle (finance/e-commerce) \cite{KaggleAPI} and 20 from OpenML \cite{Vanschoren20131, Feurer2021} (Table~\ref{tab:datasets}).
Finally, we also include MIMIC-Demo-Ext, a curated subset of the MIMIC-III Demo \cite{johnson2019mimic, Johnsonmimic2016, PhysioNet}. MIMIC-Demo-Ext contains information from the MIMIC-III Clinical Database Demo\footnote{\url{https://physionet.org/content/mimiciii-demo/1.4/}}, which is made available under the Open Database License (ODbL)\footnote{\url{https://physionet.org/content/mimiciii-demo/view-license/1.4/}}.
MIMIC-III demo comprises deidentified medical records of over 40,000 ICU patients at Beth Israel Deaconess Medical Center (2001–2012).
From these, we extracted and curated a small subset of 100 patients' records to focus on the detection of personal data while preserving the relational nature of the MIMIC-III demo.
The dataset ensures that the columns remain contextually linked through identifiers such as patient IDs or admission IDs. Additionally, we ensured that each column contains at least some non-empty values across its records, avoiding fully empty columns. This guarantees that every column contributes meaningful information for personal data detection without compromising the integrity or usability of the relational database schema.
The Kaggle, OpenML, and MIMIC-Demo-Ext datasets were not annotated for the detection of personal data.
For this reason, one of the authors performed a manual binary labeling (personal/non-personal) based on the definition of personal data of the GDPR
and the data context.

\begin{table}[h!]
    \centering
    \caption{Statistics of the datasets used}
    \label{tab:dataset_stats}
    \begin{tabular}{@{}l r r r@{}}
        \textbf{Dataset} & \textbf{Personal} & \textbf{Non-Personal} & \textbf{Total} \\
        \hline
        DeSSI (test set) & 3413 & 2859 & 6272 \\
        Kaggle & 155 & 91 & 246 \\
        OpenML & 82 & 176 & 258 \\
        MIMIC-Demo-Ext & 43 & 120 & 163 \\
        \hline
    \end{tabular}
\end{table}

As can be seen from Tab.~\ref{tab:dataset_stats} the number of features in the different datasets and the ratio of personal to non-personal features varies a lot from dataset to dataset. 
DeSSI is very balanced wrt. to personal vs. non-personal features and contains almost ten times as many features as the other datasets taken together.
MIMIC-Demo-Ext on the other hand is the smallest dataset and dominated by non-personal features.
However, MIMIC-Demo-Ext is the only actual real-world dataset.
DeSSI is mainly synthetic and for some of the datasets on Kaggle and OpenML it is not clear if they are truly authentic datasets or only inspired by real data.

\subsection{Experimental Procedure GPT-4o}

For our GPT-4o-based approach, we integrate information on the column in question with information from all other columns in the dataset (see Sec. ~\ref{sec:promptGPT1}). This is in contrast to e.g. CASSED which evaluates each column independently from all other columns.
Input prompts for our approach are structured to include the following information :
\begin{itemize}
    \item Title of the dataset
    \item Description of the dataset
    \item Column Name (feature to be classified)
    \item Names of other features of the dataset
    \item Ten most frequent values found in the column
\end{itemize}
This structured input allows GPT-4o to focus on the immediate context relevant to the column being analyzed, thereby reducing potential overload from extraneous data. The output of each column is a binary classification indicating whether it contains personal data ($True$) or not ($False$).

\subsubsection{CRSRF Framework}

The CRSRF (Capacity and Role, Statement, Reason, Format)\cite{Yang2023}, detailed in appendix~\ref{sec:CRSRF}, is designed to enhance the effectiveness of prompt-based classification tasks in machine learning models, particularly in identifying semantic relationships within datasets. This framework emphasizes the necessity of clarifying the model's role in the classification process, articulating specific objectives, and outlining expected output formats. Using this structured approach, the model can better navigate and understand complex data inputs, leading to improved classification accuracy.

\subsubsection{Prompting Structure}

The design of the prompt is crucial, as it significantly impacts the performance of the model. The prompt is structured into three main components:

\begin{enumerate}
    \item \textbf{Initial Prompt}: This component introduces the task to GPT according to the CRSRF framework. It emphasizes the importance of the task and outlines how the results should be outputted.
    
    \item \textbf{Example Prompt}: This prompt consists of an example question and answer, providing a single instance of the task (one-shot learning) to demonstrate the expected output format.
    
    \item \textbf{Data Prompt}: This component contains the specific column to be classified, along with meta-information regarding the dataset, such as the title and description.
\end{enumerate}

The complete structure of the prompt provided to the GPT API is illustrated as follows:

\noindent\hrulefill
\begin{verbatim}
conversation = [
    {"role": "system", "content": initial_prompt},
    {"role": "user", "content": example_prompt},
    {"role": "assistant", "content": example_answer},
    {"role": "user", "content": data_prompt}
    ]
\end{verbatim}
\noindent\hrulefill

In this structure, roles are defined to guide the model regarding the following entities: 

\begin{itemize}
    \item The \texttt{"system"} provides task instructions to the model.
    \item The \texttt{"user"} reflects input data that necessitates classification by the GPT model.
    \item The \texttt{"assistant"} projects the anticipated model response.
\end{itemize}
An example of a final prompt is presented in Sec. ~\ref{sec:promptGPT1}.

For each part of the prompt, a random seed is set; however, it is essential to note that this does not guarantee reproducibility of the responses of the GPT-4o model. 

\subsection{Benchmark Model Selection}
\label{subsec:presidio_cassed}
We selected Presidio \cite{Presidio} and CASSED \cite{Kuzina2023} as benchmark models against which we compare our GPT-4o-based approach. 
We selected CASSED because it is a recent model that outperformed Sherlock and SIMON in a previous comparison \cite{Kuzina2023}. We also selected Presidio as a baseline due to its frequent use and in general good performance. We did not include Sherlock and SIMON as they showed significantly weaker performance than the CASSED model in the aforementioned comparison.

To be usable in our experiments, we needed to make some adjustments to the models.
\subsubsection{Adjustments Presidio}

Presidio contains different modules.
The Presidio Analyzer module can detect PII information in textual documents, while Presidio Structured is designed to recognize PII data in tabular datasets.
We tested both modules using different approaches. For every dataset, the Presidio Analyzer outperformed the Presidio Structured module.
Consequently, we only present results for the Presidio Analyzer module.

For the Presidio Analyzer module, tabular datasets must be converted into textual data. We tested two strategies:
\begin{itemize}
    \item Columnwise, where all values from a single column, along with the column name, are provided to the model.
    \item Rowwise, where all values from a single row are transmitted together.
\end{itemize}
Presidio then predicts all detectable entities for each column or row. Next, the predicted entities are mapped to personal or non-personal (see Sec.~\ref{subsec:Presidio_class_mapping}).

To optimize the prediction accuracy for the Presidio Analyzer module, two thresholds are implemented.
The first threshold defines the minimum number of times an entity must be detected in a column to be considered valid.
The second threshold is a minimum for the confidence score of the entity, which indicates how confident Presidio is that the entity is detected correctly. As some entities that Presidio can detect are not necessarily related to a person, the predictions have to be mapped to personal and non-personal. For each dataset, the best presidio analyzer approach was used for comparing Presidio's performance against the other models in the experiments.

\subsubsection{Adjustments to CASSED}
The CASSED model was used with the settings described in the original work \cite{Kuzina2023}. Predictions are made for each column of a dataset using a column-wise approach. The input to the model is constructed for each column by combining the column header with multiple cell values from the same column, separated by delimiters('.',  ',')\cite{Kuzina2023}.

CASSED originally is able to detect 20 different classes. 
To adapt CASSED from multiclass to binary classification, we modified and retrained the model by mapping the original multiclass labels to binary labels using the label mapping described in (Sec.~\ref{subsec:CASSED_class_mapping}). 
We trained CASSED using the train and validation split of the DeSSI dataset.
Afterward, the model was evaluated on the test set of DeSSI and all other datasets (which served only as test data).  The Adam optimizer was used for fine-tuning with a learning rate of \(5 \times 10^{-5}\). The model was trained with a mini-batch size of 16 for 20 epochs, following the procedure outlined in \cite{Kuzina2023}.
In the original work of the CASSED model, the output of the DistilBERT model was combined with some rule-based heuristics, regular expressions, and lookup tables \cite{Kuzina2023}.
The publicly available CASSED model published on Github \cite{CASSEDgithub} does not contain these enhancements, so only the Transformer model was used in this work.
In the results of CASSED's original paper\cite{Kuzina2023}, the public model is only slightly worse than the enhanced version.
Consequently, the use of the publicly available model instead of the best implementation should not result in significant performance loss.

\subsection{Evaluation Metrics}  
In assessing the models' detection capabilities, we utilize the following metrics:  
\begin{itemize}  
\item \textbf{Macro F1 Score}: it assigns equal weight to all classes regardless of frequency by computing the F1 score for each class separately before computing the unweighted mean. As it assigns equal weight to all classes, including the less frequent ones, 
it is particularly useful when assessing model performance on class-imbalanced datasets.

\item \textbf{Micro F1 Score:} it calculates the F1 score globally by adding up the total number of true positives, false positives, and false negatives across all the classes before precision and recall calculation. It uses one score to measure the model's performance over all instances combined. Micro F1 is dominated by majority class performance in class-imbalanced data and therefore doesn't work well for evaluating performance in minority classes.

\item \textbf{Balanced Accuracy:} It computes the average of recall for all classes, ensuring that performance is fairly assessed even in cases of class imbalance. This metric provides a better estimate when certain categories occur less frequently in the test set.
\end{itemize}  

The models are evaluated using macro, micro F1 score, and Balanced Accuracy--valuable metrics for assessing a model's performance in multiclass classification problems, allowing a balanced view of precision and recall across classes.

\section{Experimental Results}

In this section, we compare the performance of Presidio, CASSED, and our GPT-4o-based approach on the aforementioned data sets.

\subsection{F1 Scores and Balanced Accuracy}

Our main target is the binary distinction between personal and non-personal data. 
To achieve this, we adapted CASSED to such a binary classification task (see Sec. ~\ref{subsec:presidio_cassed}).
In the case of Presidio, we performed a multiclass classification and then mapped the detected classes to either personal or non-personal categories (see Sec.~\ref{subsec:Presidio_class_mapping}).

\begin{table}[h]
\centering
    \caption{Performance comparison of Microsoft Presidio, CASSED, and GPT-4o across different datasets.}
    \label{tab:results_summary}
\begin{tabular}{l l c c c}
        \hline
        \textbf{Dataset} & \textbf{Metric} & \textbf{Presidio} & \textbf{CASSED} & \textbf{GPT-4o} \\
        \hline
        \multirow{3}{*}{DeSSI} 
            & Macro F1 & 0.793 & \textbf{0.996} & 0.766 \\
            & Micro F1 & 0.794 & \textbf{0.996} & 0.772 \\
            & Balanced Acc. & 0.791 & \textbf{0.996} & 0.764 \\
        \hline
        \multirow{3}{*}{Kaggle} 
            & Macro F1 & 0.293 & 0.349 & \textbf{0.902} \\
            & Micro F1 & 0.297 & 0.386  & \textbf{0.907} \\
            & Balanced Acc. & 0.299 & 0.481 & \textbf{0.910} \\
       \hline
        \multirow{3}{*}{OpenML} 
            & Macro F1 & 0.684 & 0.501 & \textbf{0.964} \\
            & Micro F1 & 0.733 & 0.686 & \textbf{0.969} \\
            & Balanced Acc. & 0.518 & 0.535 & \textbf{0.968} \\
        \hline
        \multirow{3}{*}{MIMIC-Demo-Ext} 
            & Macro F1 & 0.662 & 0.724 & \textbf{0.829} \\
            & Micro F1 & 0.730 & 0.798 & \textbf{0.859} \\
            & Balanced Acc. & 0.667 & 0.713 & \textbf{0.852} \\
        \hline
        \multirow{3}{*}{\textbf{Average}} 
            & \textbf{Macro F1} & 0.608 & 0.643 & \textbf{0.865} \\
            & \textbf{Micro F1} & 0.639 & 0.717 & \textbf{0.877} \\
            & \textbf{Balanced Acc.} & 0.569 & 0.681 & \textbf{0.874} \\
        \hline
    \end{tabular}
\end{table}
 
As can be seen from Tab.~\ref{tab:results_summary}, on the DeSSI dataset, CASSED performed nearly perfectly, with Macro F1, Micro F1, and Balanced Accuracy all reaching 0.996. 
Microsoft Presidio followed with a Macro F1 of 0.794, while our GPT-4o-based approach performed slightly lower at 0.766. 

Across the Kaggle and OpenML datasets, GPT-4o achieved the highest performance, with a Macro F1 score of 0.902 on Kaggle and 0.964 on OpenML. 
The performance of CASSED dropped drastically for these two datasets to Macro F1 scores of 0.349 and 0.501, respectively.
Similarly, Presidio also only achieved Macro F1 scores of 0.293 and 0.684 on these datasets.
The differences in performance were less pronounced for the MIMIC-Demo-Ext dataset.
Here again, our GPT-4o-based approach leads with a Macro F1 score of 0.865.
CASSED and Presidio achieve 0.724 and 0.662, respectively.
Hence, CASSED shows clearly superior performance on DeSSI, a dataset on which it was developed.
For all other datasets, our GPT-4o-based approach is better.
Presidio shows comparable performance to CASSED except for the evaluation on DeSSI.
This is also visible when averaging scores over all datasets.
 Here, our GPT-4o-based approach shows with an averaged Macro F1 score of 0.865 clearly superior performance to CASSED (0.643) and Presidio (0.608).
The other measures (Micro F1 and Balanced Accuracy) show a similar behavior. 

\subsection{Analysis of False Negatives and False Positives}

\begin{figure}[h]
    \centering
    \includegraphics[scale=0.52]{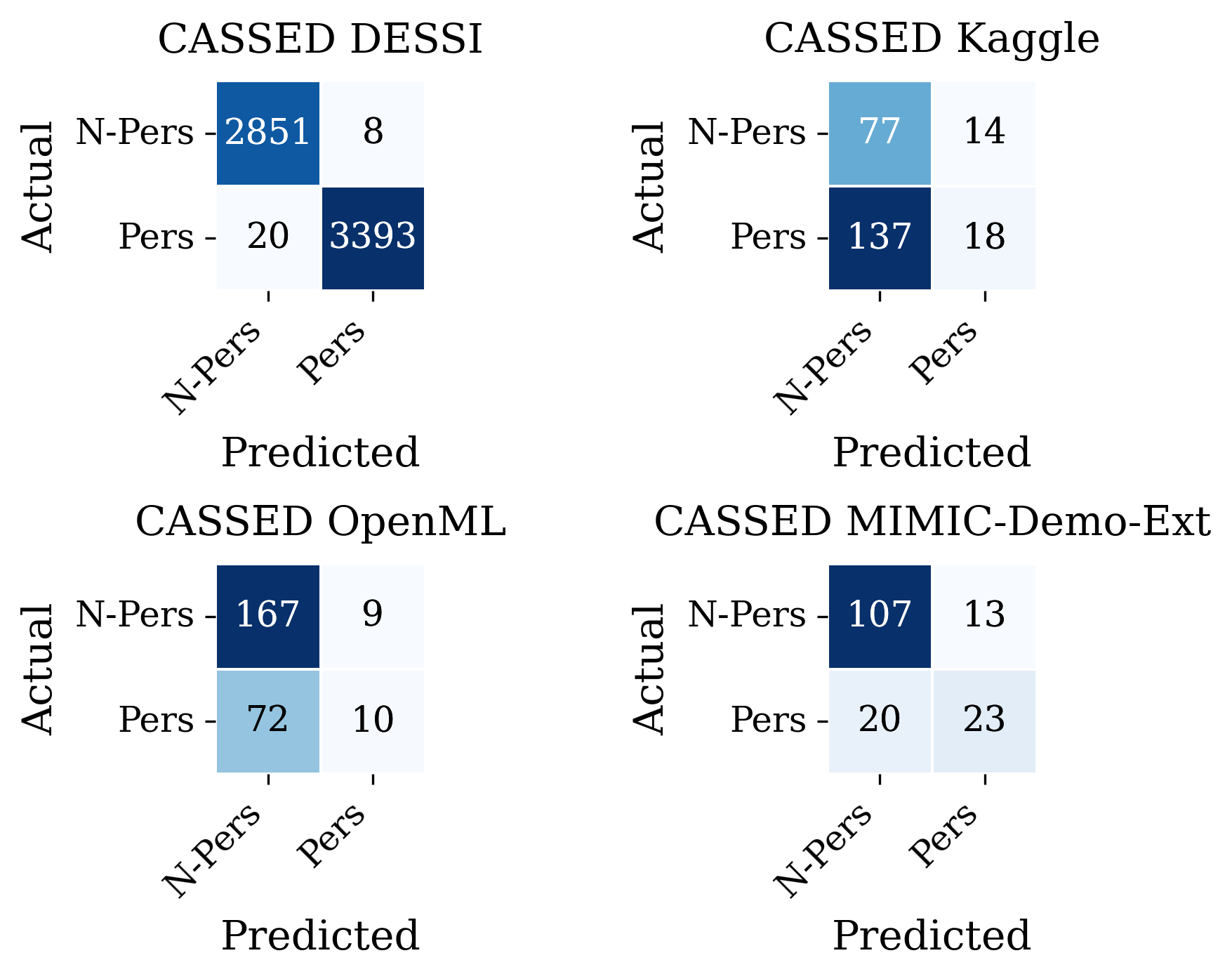}
    \Description{Confusion matrices illustrating the performance of CASSED on the DESSI, Kaggle, OpenML, and MIMIC-Demo-Ext datasets. Each matrix shows the classification results for personal and non-personal data, including true positives, false positives, true negatives, and false negatives.}
    \caption{Confusion matrices for CASSED model performance across datasets: DESSI, Kaggle, OpenML, and MIMIC-Demo-Ext. Each matrix shows the classification results for personal (P) and non-personal (N-pers) data categories.} 
    \label{fig:CM_cassed}
\end{figure}

\begin{figure}[h]
    \centering
    \includegraphics[scale=0.52]{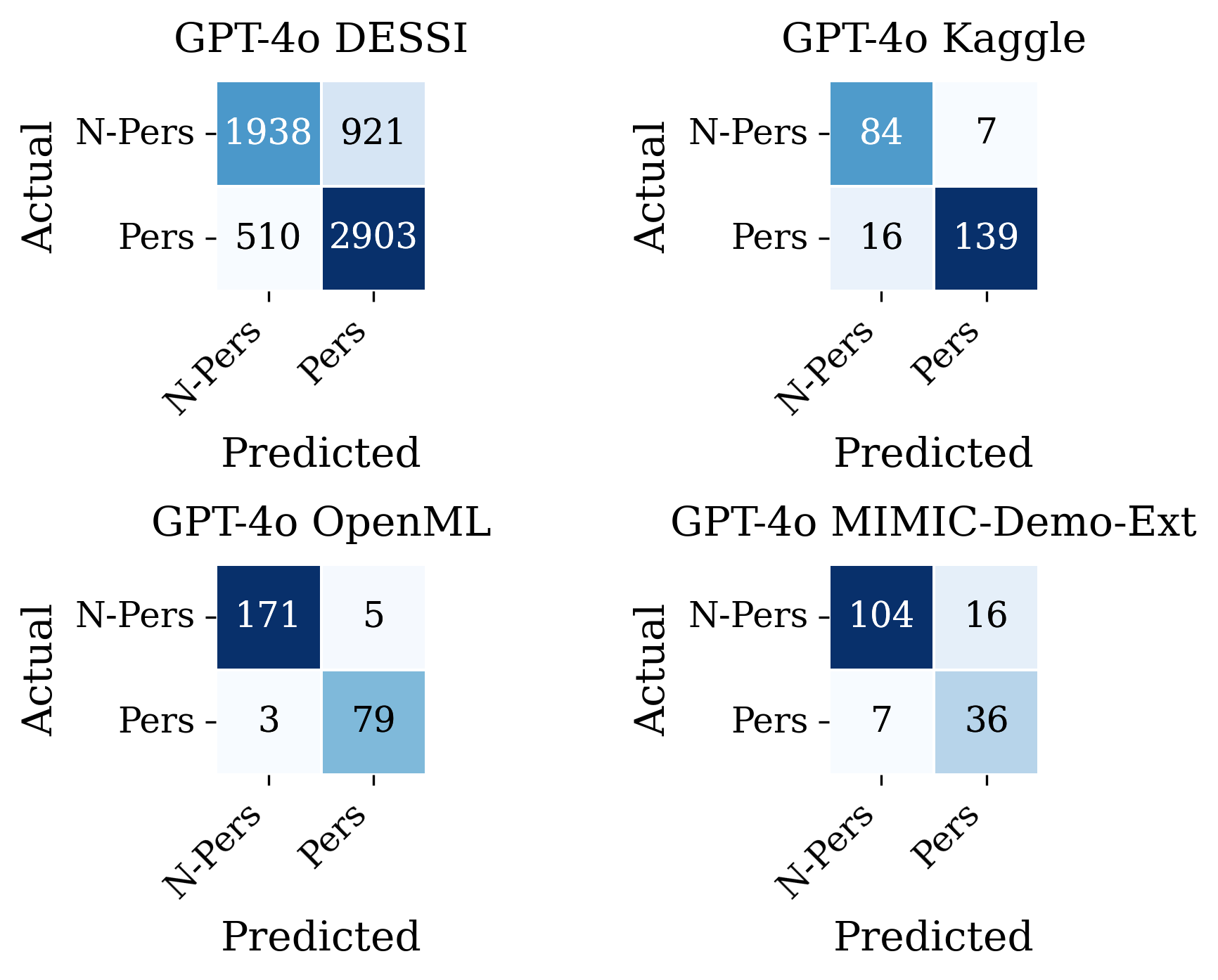}
    \Description{Confusion matrices illustrating the performance of our GPT-4o-based model on the DESSI, Kaggle, OpenML, and MIMIC-Demo-Ext datasets. Each matrix displays true positives, false positives, true negatives, and false negatives for personal and non-personal data classification.}
    \caption{Confusion matrices for our GPT-4o-based model performance across four datasets: DESSI, Kaggle, OpenML, and MIMIC-Demo-Ext. Each matrix shows the classification results for personal (P) and non-personal (N-pers) data categories. }
    \label{fig:CM_gpt40}
\end{figure}

\begin{figure}[h]
    \centering
\includegraphics[scale=0.52]{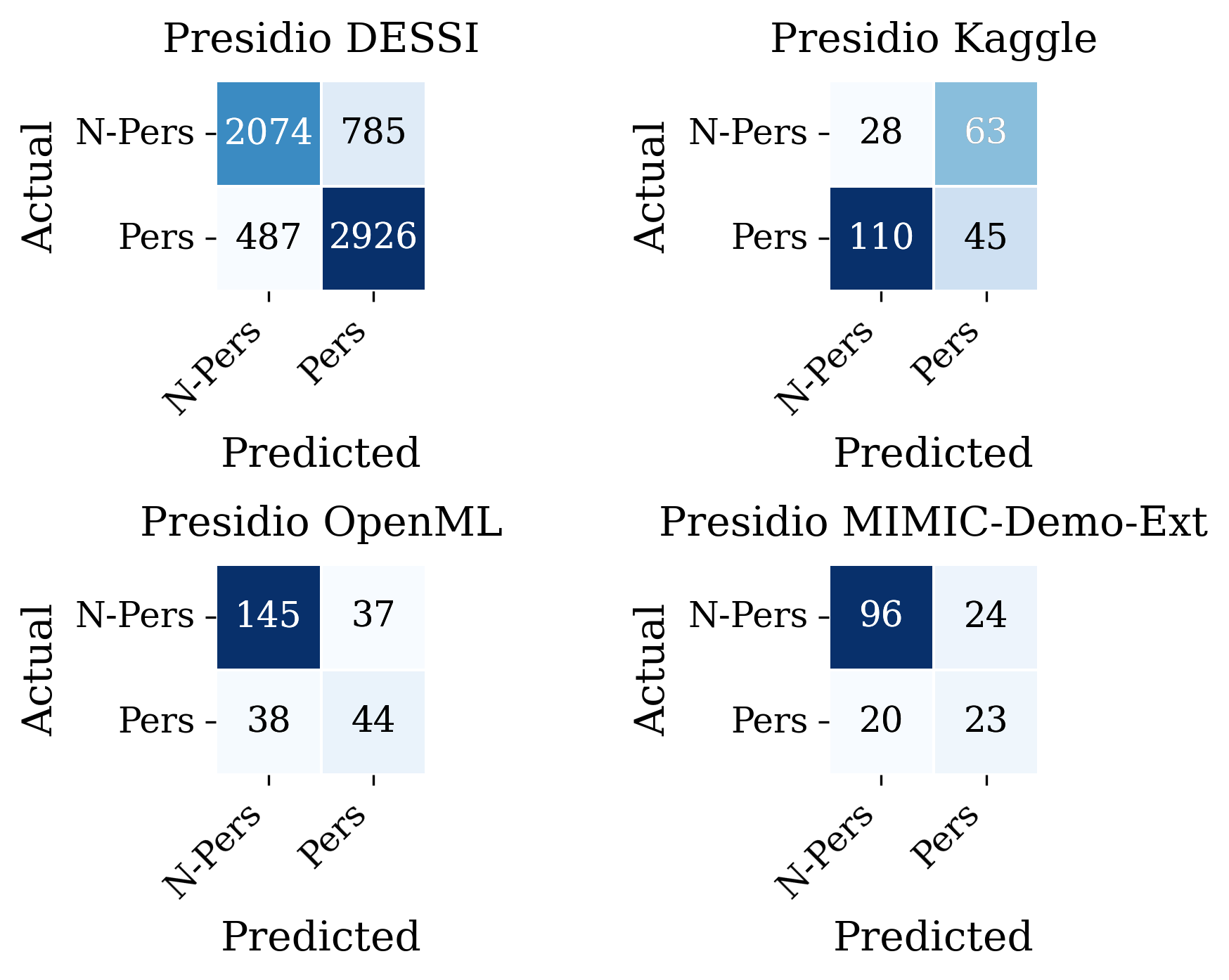}
    \Description{Four confusion matrices illustrating Presidio model performance on DESSI, Kaggle, OpenML, and MIMIC-Demo-Ext datasets. Each matrix shows the classification results for personal and non-personal data, including true positives, false positives, true negatives, and false negatives.} 
    \caption{Confusion matrices for Presidio model performance across datasets: DESSI, Kaggle, OpenML, and MIMIC-Demo-Ext. Each matrix shows the classification results for personal (P) and non-personal (N-pers) data categories.} 
    \label{fig:CM_presidio}
\end{figure}

A model's ability to prevent false negatives (FN) has high practical relevance, as failing to detect personal data can pose serious GDPR compliance risks. 
When looking at the confusion matrices for all three approaches and all 4 datasets (Fig.~\ref{fig:CM_cassed}, \ref{fig:CM_gpt40}, and \ref{fig:CM_presidio}) one can observe that CASSED performs near perfect on DeSSI yet shows frequent false negatives (personal-related data not detected as such) for the Kaggle and OpenML datasets. This is also visible for the MIMIC-Demo-Ext dataset but less pronounced.
The performance of our GPT-4o-based approach is more balanced for all datasets other than DeSSI.
For DeSSI it shows a notable tendency for false positives (non-person-related detected as person-related).
The performance of Presidio is the most balanced yet overall inferior.

Hence, the risk of false negatives is notably smaller for our GPT-4o-based approach compared to CASSED and Presidio for the real-world datasets (Kaggle, OpenML, MIMIC-Demo-Ext) but clearly inferior to CASSED on the synthetic data (DeSSI).
Nevertheless, in light of the costs potentially involved, the FN rates on the real-world data are currently still too high for practical applications.  

\subsection{Analysis of Selected Examples}
\renewcommand{\arraystretch}{1.2}
\begin{table}[h]
\caption{Examples in DeSSI Data}
    \label{tab:dessi_false_negatives}
\begin{tabular}{@{}l@{\hspace{5pt}} l@{\hspace{5pt}} l@{\hspace{5pt}} l@{\hspace{5pt}}}
        \textbf{Features} & \textbf{oib.1} & \textbf{accountid.3} & \textbf{kzwwskxzgjgc} \\
        \midrule
        \multirow{3}{*}{Values} 
            &ZZ 563634 T &  Jasper & g-h@hotmail.com \\
            & ZZ 140837 T & vocativ &s.draksic0@ribaric.com \\
            &  ZZ 88 70 27 T	& kinzo berlin	 & g-dominguez@wu.net \\[0.25em]
        \multirow{1}{*}{Context} 
            & Synthetic data & Synthetic data & Synthetic data \\
        \multirow{1}{*}{True Label} 
            & Personal & Personal & Personal \\
        \multirow{1}{*}{CASSED} 
            & Personal & Personal & Personal \\
        \multirow{1}{*}{GPT-4o} 
            & Non-personal & Non-personal & Non-personal\\
        \midrule
    \end{tabular}
\end{table}

\begin{table}[h]
\centering
\caption{Examples in Kaggle Data}
    \label{tab:kaggle_false_negatives}
\begin{tabular}{@{}l@{\hspace{5pt}} l@{\hspace{5pt}} l@{\hspace{5pt}} l@{\hspace{5pt}}}
        \textbf{Features} & \textbf{Cabin} & \textbf{Ticket} & \textbf{Reason Absence} \\
        \midrule
        \multirow{3}{*}{Values} 
            & C103 &  A/5 21171 & 26  \\
            & C123 & STON/O2,3101 &0  \\
            &  E46 & 374910 & 19 \\
        \multirow{1}{*}{Context} 
            & Titanic data & Titanic data & Absenteeism \\
        \multirow{1}{*}{True Label} 
            & Personal & Personal & Personal \\
        \multirow{1}{*}{CASSED} 
            & Non-personal & Non-personal & Non-personal \\
        \multirow{1}{*}{GPT-4o} 
            & Personal & Personal & Personal \\
        \midrule
    \end{tabular}
\end{table}

\begin{table}[h]
\centering
\caption{Examples in OpenML Data}
    \label{tab:openml_false_negatives}
\begin{tabular}{@{}l@{\hspace{5pt}} l@{\hspace{5pt}} l@{\hspace{5pt}} l@{\hspace{5pt}}}
        \textbf{Features} & \textbf{Email Address} & \textbf{Location} & \textbf{Customer City} \\
        \midrule
        \multirow{3}{*}{Values} 
           &alexandra@example.org & Rebeccachester & sao bernardo \\	
            & holland@example.com & sao paolo & niteroi  \\
            & elizabeth31@example.net & Port Deborah & campinas \\
        \multirow{1}{*}{Context} 
            & Customer data & Customer data & Customer data \\
        \multirow{1}{*}{True Label} 
            & Personal & Personal & Personal \\
        \multirow{1}{*}{CASSED} 
            & Non-personal & Non-personal & Non-personal \\
        \multirow{1}{*}{GPT-4o} 
            & Personal & Personal & Personal \\
        \midrule
    \end{tabular}
\end{table}

\begin{table}[h]
\centering
\caption{Examples in MIMIC-Demo-Ext Data}
    \label{tab:mimic_false_negatives}
\begin{tabular}{@{}l@{\hspace{5pt}} l@{\hspace{5pt}} l@{\hspace{5pt}} l@{\hspace{5pt}}}
        \textbf{Features} & \textbf{marital\_status} & \textbf{discharge\_location} & \textbf{last\_careunit} \\
        \midrule
        \multirow{3}{*}{Values} 
            &MARRIED &  HOME & MICU \\
            & DIVORCED & SNF &CCU \\
            &  SEPARATED & DEAD/EXPIRED & TSICU \\
        \multirow{1}{*}{Context} 
            & Medical data & Medical data & Medical data \\
        \multirow{1}{*}{True Label} 
            & Personal & Personal & Personal \\
        \multirow{1}{*}{CASSED} 
            & Non-personal & Non-personal & Non-personal \\
        \multirow{1}{*}{GPT-4o} 
            & Personal & Personal & Personal \\
        \midrule
    \end{tabular}
\end{table}

As we could see above, the performance of the different models varied significantly depending on the dataset.
We will now have a closer look at detailed results for individual features in the different datasets.
As Presidio showed inferior performance we limit this analysis to CASSED and our GPT-4o-based approach.

Tab.~\ref{tab:dessi_false_negatives} shows some examples from the DeSSI dataset which CASSED successfully recognized as personal and our GPT-4o-based approach failed.
CASSED seems to have learned the corresponding relations quite well and is in the case of the last feature able to detect the e-mail addresses. 
Our GPT-4o-based approach seems to be confused by the rather uninformative feature name.
CASSED seems also to know (via BERT or have learned from other examples) that "OIB" is a permanent national identification number of every Croatian citizen and correctly identifies it as personal. 
Again, our GPT-4o-based approach struggles here, presumably because the names of the other features in the dataset - contextual cues our approach integrates - do not give enough hints or because of insufficient representation of Croatia in the training data of GPT-4o.
The situation for "accountid" is similar.

When looking at the examples for the Kaggle data in Tab.~\ref{tab:kaggle_false_negatives} we see that our GPT-4o-based approach was able to use the context given by the dataset description and the names of the other features to correctly identify the personal information. 
Here CASSED struggled as some context is required to infer that "Cabin", "Ticket" and "Reason Absence" might reveal personal information. 
Most likely for similar reasons CASSED also misclassified domain-specific attributes such as "Workclass" and "Income".

The examples where CASSED failed in the OpenML data in Tab.~\ref{tab:openml_false_negatives} are a bit surprising.
"Email Address" and "Customer City" should be identifiable as personal also without context.
Possibly a lack of variation in the synthetically created training data of CASSED prevents it from detecting e-mail addresses containing "example" in the domain.
It is also a possibility that these types of e-mail addresses were explicitly labeled as non-personal in the training data.
CASSED also missed crucial identifiers like "User ID". 
Conversely, it produced false positives, incorrectly flagging anonymized "Address" and "Postcode" fields as sensitive, and mislabeling generic terms like "Referee" and "Species" as personal data. 
These errors highlight CASSED's problems in dealing with domain-specific scenarios where contextual interpretation is crucial for accurate classification.
Yet our GPT-4o-based approach did also not perform perfectly on the OpenML data. 
In some instances, it misclassified fields like "userid" and "customer\_id.13" as non-personal, likely due to their generic naming conventions.

The mistakes of CASSED we see on the MIMIC-Demo-Ext dataset (compare Tab.~\ref{tab:mimic_false_negatives}) might be due to its special nature: medical data. 
It is possible that medical data was not represented sufficiently in the training data of CASSED. Nevertheless, it is surprising that it did not recognize a feature as "marital\_status" correctly. This hints at other reasons than the unfamiliarity with medical data for its poor performance.

\section{Discussion}

The results showed a very strong performance of CASSED on the DeSSI dataset (compare Tab.~\ref{tab:results_summary}).
On the other hand, CASSED's performance decreased notably when evaluated on the MIMIC-Demo-Ext dataset, and it performed rather poorly on Kaggle and OpenML.
Based on our current analysis it is difficult to determine the reasons for this.
One possible explanation is overfitting of CASSED on DeSSI, the data set that was developed in conjunction with CASSED.

This could be due to DeSSI representing only a narrow subset of the true variation in personal data or a consequence of the train/dev/test split.
Notably, the authors do not specify whether precautions were taken to ensure that features from the same dataset were not distributed across different splits. If this was not accounted for, the system might have leveraged information from the training set when processing the test set, as these features cannot be assumed to be entirely independent.
It could also stem from the synthetic generation of features, where the same underlying patterns may have been used for features appearing in both the training and test splits.
Another possible explanation is that the individual features in the dataset are largely independent, making them highly compatible with CASSED's core assumptions.
Based on our more detailed analysis of some examples in Tab:~\ref{tab:kaggle_false_negatives}-\ref{tab:mimic_false_negatives} we suspect at least some overfitting of CASSED on DeSSI.
Otherwise, it is hard to explain why it would have difficulties recognizing "Email Address", "Customer City" and "marital\_status" correctly.
Another possibility could be a misalignment in the annotations we used and those used for DeSSI.
However, when looking at their classes and our mapping (compare Tab:~\ref{tab:dessi_personal_nonpersonal}) it is unlikely that this explains the results.

The poor performance of CASSED and Presidio on the Kaggle and OpenML datasets could also be explained by the contextual information required to deal with these datasets.
If features like "Cabin", "Ticket", and "Location" contain personal information, it depends, in general, on context.
This helps explain the superior performance of our GPT-4o-based approach on these datasets.
Another reason might be that these well-known and widely distributed datasets were contained in GPT-4o's training data.
This might give GPT-4o an advantage even though they are, to our knowledge, not available with annotations for the detection of personal data (this annotation is our own).
Finally, on the MIMIC-Demo-Ext dataset, the differences in performance between the different models were less striking.
Here our GPT-4o-based approach obtained inferior results to those on Kaggle and OpenML.
Nevertheless, it quite clearly outperformed CASSED and Presidio.
One reason for this could be that medical data was not well represented in DeSSI, CASSED's training data.  
This would be rather unfortunate as personal data detection is a very important topic in the medical domain.
GPT-4o, with its vast amount of training data, might hence be better able to cope with this.
Another reason might be the benefits of contextual information for this dataset.
From the examples we analyzed in Tab:~\ref{tab:mimic_false_negatives} it is difficult to make conclusions about this.

Additionally, the datasets we investigated can also be split across another dimension: real-world (Kaggle, OpenML, MIMIC-Demo-Ext) vs. synthetic (DeSSI).
Here, the conclusion might be that the DeSSI dataset does not represent the real world sufficiently well, which leads to the very clear drop in the performance of CASSED when applied to real-world data.
However, when looking at the results, it has to be kept in mind that DeSSI alone contains roughly 10 times as many features as the other datasets taken together.
More thorough conclusions will require the analysis of more and larger real-world datasets.
However, the requirements for data protection complicate the access to such datasets.

\subsection{Limitations and Future Research}

We demonstrated that LLMs, particularly when incorporating context, can enhance the detection of personal data, particularly in real-world datasets. However, current performance is still insufficient to minimize the risk of personal data disclosure and ensure full GDPR compliance. 
As main avenues for further improvement, we see:

\begin{itemize}

\item Dataset Diversity: The study has highlighted some potential limitations of current methods resulting from the use of synthetic data. It seems likely that DeSSI, the only large-scale dataset for personal data detection, is not diverse enough to cover real-world variations. Additional large and diverse real-world datasets are needed to make further progress.

\item Influence of context: We could give some hints that the use of contextual information in our GPT-4o-based model was beneficial yet a more detailed analysis is needed to better assess what role context plays and how it can be most effectively used.

\item  Privacy Constraints: Since GPT-4o operates as an online service, its reliance on cloud-based processing raises significant privacy concerns. Transmitting personal data to external servers poses compliance risks under data protection regulations. Future research needs to explore secure on-premise solutions to overcome this. 

\item Computational Demand: GPT-4o is a very powerful yet also very computationally demanding model requiring orders of magnitude more computational resources than Presidio or the BERT-based CASSED. Hence, in addition to the need to find on-premise solutions also much smaller LLMs need to be investigated. 

\item  Hybrid models: It can be expected that integrating ideas from all three models (rule-based approaches, classical machine learning, and LLMs) will help to further improve results.

\item False Negatives: Concerning the high priority of not accidentally revealing personal information, adaptations to the models need to be made to better control false negatives.
\end{itemize}

\section{Conclusion}
Despite significant performance differences across datasets, we conclude that our GPT-4o-based approach is the most effective model for detecting personal data in structured datasets.
It demonstrated strong performance on synthetic data in DeSSI, outperformed CASSED and Presidio on medical data in MIMIC-Demo-Ext, and clearly surpassed both approaches on Kaggle and OpenML—likely due to its ability to leverage contextual information.

However, this high performance comes at the cost of substantial computational demands. Future research should explore whether smaller, locally running LLMs can achieve comparable results with lower resource requirements.

Additionally, further investigation is needed to determine whether CASSED’s exceptional performance on DeSSI reflects genuine model capabilities or is merely an artifact of overfitting to this artificial dataset. A more comprehensive analysis, along with additional real-world datasets, will be crucial for a robust evaluation of these models.

Overall, we have demonstrated that high-performance personal data detection in structured datasets is achievable. However, further advancements are necessary to minimize the risk of unintended data exposure and ensure these methods meet acceptable privacy standards.

\begin{acks}
This work was conducted in the context of the project KI-Allianz BW: Datenplattform funded by Ministerium für Wirtschaft, Arbeit und Tourismus Baden-Württemberg. 
\end{acks}

\bibliographystyle{ACM-Reference-Format}
\bibliography{PIIDetection_references}

\appendix

\section{Prompt to GPT for Datasets with Context}
\label{sec:promptGPT1}

\begin{tcolorbox}[colback=gray!10, colframe=violet!80, title=Prompt to GPT for Datasets with Context]

\textbf{Initial Prompt:}  
As a classifier of person-related data in tabular datasets, your task is to analyze the provided columns (each containing up to ten distinct values) and determine whether they contain information that originates from or relates to a person, even if it is not directly identifiable.  
Detecting person-related information helps ensure compliance with data protection regulations and safeguards individuals’ privacy and security.  
Output your results in a dictionary format with a boolean indicating if the column contains person-related data or not.

\textbf{Example Prompt:}

You can use the following example as a guideline:  
Classify the following column with careful consideration of the dataset description:  

\textbf{Dataset:}

\textbf{Title:} ‘Test Dataset‘ 

\textbf{Description:} ‘This dataset was used for a linear regression.‘  

\textbf{Features:} \texttt{['first\_name\_en\_10', 'last\_name\_en\_10', 'email\_en\_10', 'phone\_number',  
'address\_en\_10', 'city\_en\_10', 'country\_en\_10', 'date', 'target']}  

\textbf{Column of the dataset to classify:}  
\texttt{'first\_name\_en\_10': ['Tom', 'Walter', 'Mia', 'Lena', 'John', 'Jack', 'Felice', 'Anna', 'Lukas', 'Will']}  

Does this column, in the context of the dataset, contain information relating to a natural person?  

\textbf{Example Answer:}  
\texttt{\{'first\_name\_en\_10': true\}}

\textbf{Data Prompt:}  
Classify the following column with careful consideration of the dataset description.  

\textbf{Dataset:} 
\textbf{Title:} \textit{Absenteeism at Work} 

\textbf{Description:} Context - The database was created with records of absenteeism at work from July 2007 to July 2010 at a courier company in Brazil.  

\textbf{Features:} \texttt{Index(['ID', 'Reason for absence', 'Month of absence', 'Day of the week', 'Seasons',  
'Transportation expense', 'Distance from Residence to Work', 'Service time', 'Age',  
'Work load Average/day', 'Hit target', 'Disciplinary failure', 'Education', 'Son',  
'Social drinker', 'Social smoker', 'Pet', 'Weight', 'Height', 'Body mass index',  
'Absenteeism time in hours'], dtype='object')}

\textbf{Column of the dataset to classify:}  \texttt{'ID': [3, 20, 28, 11, 15, 34, 10, 33, 14, 36]}

Does this column, in the context of the dataset, contain information relating to a natural person?  

\textit{Note: The description of the dataset has been shortened for better readability.}
\end{tcolorbox}

\section{CRSRF Framework for Constructing Prompts} \label{sec:CRSRF}

The CRSRF framework provides a structured approach for designing prompts that effectively guide large language models (LLMs) in detecting and safeguarding personal information within archives. It consists of the following key components:

\begin{itemize}
    \item \textbf{Capacity and Role}: This element establishes the LLM’s task by defining its function as a detector and protector of personal information within text-based archives. The prompt may begin with:  
    \textit{“As a comprehensive identifier of personal information within text-based archives…”}  
    
    \item \textbf{Statement}: This defines the specific objective, explicitly stating the types of personal information the LLM should identify. A sample statement could be:  
    \textit{“Search for and flag any occurrences of personal names, unique identification codes such as identity card numbers or passport numbers, telephone numbers, home addresses, and mentions of family members…”}  
    
    \item \textbf{Reason}: This section provides justification for the task, emphasizing the significance of protecting personal data. A well-structured reason may be:  
    \textit{“These details, if exposed, can compromise an individual’s privacy and security. It is crucial to identify them to ensure the confidentiality and integrity of the archived documents.”}  
    
    \item \textbf{Format}: This specifies the preferred output format for the extracted information, ensuring structured and clear presentation. Given the nature of the data, a list format is recommended:  
    \textit{“Present the identified personal information in a list format, with categories such as ‘Name,’ ‘Identification Code,’ ‘Telephone Number,’ ‘Address,’ and ‘Family Members’ as keys.”}  
\end{itemize}

\section{Mapping of Classes}

\subsection{Mapping for CASSED}
\label{subsec:CASSED_class_mapping}
For the DeSSI labels, columns were labeled as personal if at least one entity belonged to the personal-related category (Table ~\ref{tab:dessi_personal_nonpersonal})

\begin{table}[ht]
\centering
\caption{Classification of entities in the DeSSI dataset}
\begin{tabular}{l l}
\hline
\textbf{Personal} & \textbf{Non-Personal} \\ \hline
Phone number & Other data \\ 
Address & Organization \\ 
Person & GPE \\ 
Email & SWIFT/BIC \\ 
NIN & Geolocation \\ 
Date & \\ 
Passport & \\ 
CCN & \\ 
ID Card & \\ 
Sexuality & \\ 
Gender & \\ 
Nationality & \\ 
Race & \\ 
Religion & \\ 
IBAN & \\ \hline
\end{tabular}
\label{tab:dessi_personal_nonpersonal}
\end{table}

\subsection{Mapping for Presidio}
\label{subsec:Presidio_class_mapping}
The classification follows the logic that direct identifiers and sensitive attributes related to individuals fall under personal, while business-related and general references are non-personal unless they reveal individual identity. This table provides an overview of the PII entities that Presidio can detect using its predefined recognizers.

\begin{table}[h]
\centering
\caption{Classification of MS Presidio Recognizers}
\label{tab:presidio_identifiers}
\begin{tabular}{l l}
\hline
\textbf{Personal} & \textbf{Non-Personal} \\
\hline
CREDIT\_CARD & DATE\_TIME \\
CRYPTO & IP\_ADDRESS \\
EMAIL\_ADDRESS & LOCATION \\
IBAN\_CODE & URL \\
NRP (Passport) & AU\_ABN \\
PERSON &  AU\_ACN\\
PHONE\_NUMBER &  \\
SSN &  \\
US\_BANK\_NUMBER &  \\
US\_DRIVER\_LICENSE &  \\
US\_ITIN &  \\
US\_PASSPORT &  \\
US\_SSN &  \\
\hline
\end{tabular}
\end{table}

\section{Kaggle and OpenML Datasets}
To evaluate the performance of our methods on diverse and real-world data, we utilized datasets from two prominent platforms: Kaggle and OpenML.
\begin{table}[h!]
\caption{List of Kaggle and OpenML Datasets used.}
\label{tab:datasets}
\centering
\begin{tabular}{l}
\hline
\textbf{Kaggle Datasets} \\ \hline
Absenteeism at Work, Adult Census Income, Agriculture,\\ Bank Marketing Campaigns, Diabetes, Graduate Admission 2,\\ Indian Companies Registration Data, Indian Liver Patient Records, \\ 
London House Price, Phishing Email,\\ Pixar Movies, Student Performance, Titanic \\ \hline
\textbf{OpenML Datasets} \\ \hline
Amazon Prime Fiction, APL\_20\_24, CSM, DATASETBANK, \\company quality and valuation finance,  FitBit HeartRate,\\ HousingPrices, mango detection australia, Oilst Customers Dataset,\\ TVS Loan Default, Avocado Prices (Augmented), echoMonths,\\ fishcatch, forest fires, FOREX chfjpy minute Close,\\ iris, Marvel Movies Dataset, nyc taxi green dec 2016,\\ vowel, wine quality\\
\hline
\end{tabular}
\end{table}

\end{document}